\documentclass{article}
\usepackage{spconf,amsmath,graphicx}

\usepackage{graphicx}
\usepackage{amsmath} 
\usepackage{amsthm}
\usepackage{booktabs}
\usepackage{algorithm}
\usepackage{algorithmic}
\usepackage{multirow}
\usepackage{subfigure}
\usepackage{xcolor}

\title{Training Robust Spiking Neural Networks on Neuromorphic Data with Spatiotemporal Fragments}
\name{Haibo Shen\textsuperscript{1}, 
Yihao Luo\textsuperscript{2,1}, 
Xiang Cao\textsuperscript{3,1}, 
Liangqi Zhang\textsuperscript{1}, 
Juyu Xiao\textsuperscript{1}, 
Tianjiang Wang\textsuperscript{1}\thanks{This work was supported in part by the National Natural Science Foundation of China under Grant 61572214 and Seed Foundation of Huazhong University of Science and
Technology (2020kfyXGYJ114). (Corresponding author: Tianjiang Wang.)}}
\address{School of Huazhong University of Science and Technology\textsuperscript{1}\\
Yichang Testing Technique Research Institute\textsuperscript{2}\\
Changsha University\textsuperscript{3}}
\begin{document}
%

\maketitle

\begin{abstract}
  Neuromorphic vision sensors (event cameras) are inherently suitable for spiking
  neural networks (SNNs) and provide novel  
  neuromorphic vision data for this biomimetic model.
  Due to the spatiotemporal characteristics, 
  novel data augmentations are required to
  process the unconventional visual signals of these cameras.
  In this paper, we propose a novel Event SpatioTemporal Fragments (ESTF) augmentation
  method.
  It preserves the continuity of neuromorphic data 
  by drifting or inverting fragments of the spatiotemporal event stream to 
  simulate the disturbance of brightness variations, 
  leading to more robust spiking neural networks.
  Extensive experiments are performed on prevailing neuromorphic datasets. 
  It turns out that ESTF provides substantial improvements 
  over pure geometric transformations and 
  outperforms other event data augmentation methods.
  It is worth noting that the SNNs with ESTF achieve the 
  state-of-the-art accuracy of 83.9\%
  on the CIFAR10-DVS dataset.
\end{abstract}

\section{Introduction}

Event cameras~\cite{ATISCamera,DVSCamera} are bio-inspired 
vision sensors that operate in a completely different way from 
traditional cameras~\cite{EST,EventSurvey}. Instead of capturing images 
at a fixed rate, the cameras asynchronously measure per-pixel 
brightness changes and output a stream of events that encode 
the time, location, and sign of the brightness changes.  
Since both events and spikes are modeled from neural signals,
event cameras are inherently suitable 
for spiking neural networks (SNNs),
which are considered promising models for artificial intelligence (AI) 
and theoretical neuroscience~\cite{roy2019towards}. 

Due to the event characteristics, brightness variances 
tend to disrupt the spatiotemporal positions 
of the events, posing a great challenge to the robustness of SNNs. 
However, traditional data augmentations are fundamentally 
designed for RGB data 
and lack exploration of neuromorphic events, such as 
mixup~\cite{ZhangCDL18Mixup} 
and random erase~\cite{Zhong0KL020Erasing}.
Therefore, novel data augmentations are required to process 
the unconventional visual signals of these cameras.

In this paper, we propose a novel 
Event SpatioTemporal Fragments (ESTF) augmentation method 
to model the effects of brightness variances on neuromorphic data. 
ESTF can be simply summarized into two components, 
inverting event fragments 
in the spatiotemporal and polarity domains, 
and drift events in the spatiotemporal domain.
ESTF preserves the original spatiotemporal structure 
through transformed fragments and increases the diversity 
of neuromorphic data.
Extensive experiments are conducted on prevailing neuromorphic datasets. 
It turns out that ESTF improves the robustness of SNNs 
by simulating the effect of brightness variances on neuromorphic data. 
Furthermore, we analyze the insightful superiority of ESTF compared with
other event data augmentation methods.
It is worth noting that ESTF is super effective on 
both SNNs and convolutional neural networks~(CNNs). 
For example, the SNNs with ESTF 
yield a substantial improvement
over previous state-of-the-art results
and achieve an accuracy of 83.9\% on the CIFAR10-DVS dataset, 
which illustrates the importance of 
event spatiotemporal fragment augmentation.

In addition, while this work is related to 
EventDrop~\cite{GuSHY21} and NDA~\cite{li2022neuromorphic}, 
there are some distinct differences. 
For example, NDA is a purely global geometric transformation, 
while ESTF transforms partial event stream 
in the spatiotemporal or polarity domain.
EventDrop introduces noise by dropping events, 
resulting in poor sample continuity.
However, ESTF also considers the  
brightness variances in neuromorphic data 
and preserves the original spatiotemporal 
structure of events.

\section{Method}
\subsection{Event Generation Model}

The event generation model~\cite{EST,EventSurvey} 
is abstracted from dynamic vision sensors~\cite{DVSCamera}. 
Each pixel of the event camera responds to changes in 
its logarithmic photocurrent $L=\log(I)$. Specifically, 
in a noise-free scenario, an event $e_k = (x_k, y_k, t_k, p_k)$ is 
triggered at pixel $ X_k = (y_k, x_k)$ and at time $t_k$ as soon as the brightness 
variation $|\Delta L|$ reaches a temporal contrast threshold $C$ 
since the last event at the pixel.
The event generation model can be expressed by the following formula:
\begin{equation}\label{event_generate}
    \begin{aligned}
        \Delta L(X_k, t_k) = L(X_k, t_k) - L(X_k, t_k - \Delta t_k) = p_k C
    \end{aligned}
\end{equation}
where $C > 0$, $\Delta t_k$ is the time elapsed since the last event at
the same pixel, and the polarity $p_{k} \in\{+1,-1\}$ is the sign
of the brightness change. During a period, 
the event camera triggers event stream $\mathcal{E}$:
\begin{equation}\label{events_stream}
    \mathcal{E}=\left\{e_{k} \right\}_{k=1}^{N}=\left\{\left(X_{k}, t_{k}, p_{k}\right)\right\}_{k=1}^{N}
\end{equation}
where $N$ represents the number of events in the set $\mathcal{E}$. 

Further, inspired by previous work~\cite{EST}, an event field is used to represent 
the discrete $\mathcal{E}$ by replacing each event of the spatiotemporal stream 
with a Dirac spike, resulting in a continuous representation of events. 
For convenience, we define the complete set of transformation domains 
as $D=\{X, p, t\}$, and map the polarity $p$ from $\{-1, +1\}$ to $\{0, 1\}$.
The continuous event filed $S$ can be expressed as the following formula:
\begin{equation}\label{events_filed}
    \begin{aligned}
        S^{\mathcal{E}}(X, p, t)&=\sum_{e_{k} \in \mathcal{E}}\delta(X-X_k)\delta(t-t_k)\delta(p-p_k)\\
         &=\sum_{e_{k} \in \mathcal{E}} \prod_{i \in D} \delta\left(i-i_{k}\right)
    \end{aligned}
\end{equation}

The event generation model is shown in Fig.~\ref{fig_logA},
an event is generated each time the brightness variances reach the threshold, 
and then $|\Delta L|$ is cleared. Note that due to the single-ended 
inverting amplifier in the sensor differential circuit, 
the threshold for brightness changes in the circuit is one sign different 
from the threshold in the conceptual model. 
In addition, the short segment following the firing event represents 
the corresponding refractory period in biological neurons.


\subsection{Motivation}

\begin{figure}[t]
    \centering
    \subfigure[Event generation model.]{\includegraphics[width=0.495\linewidth]{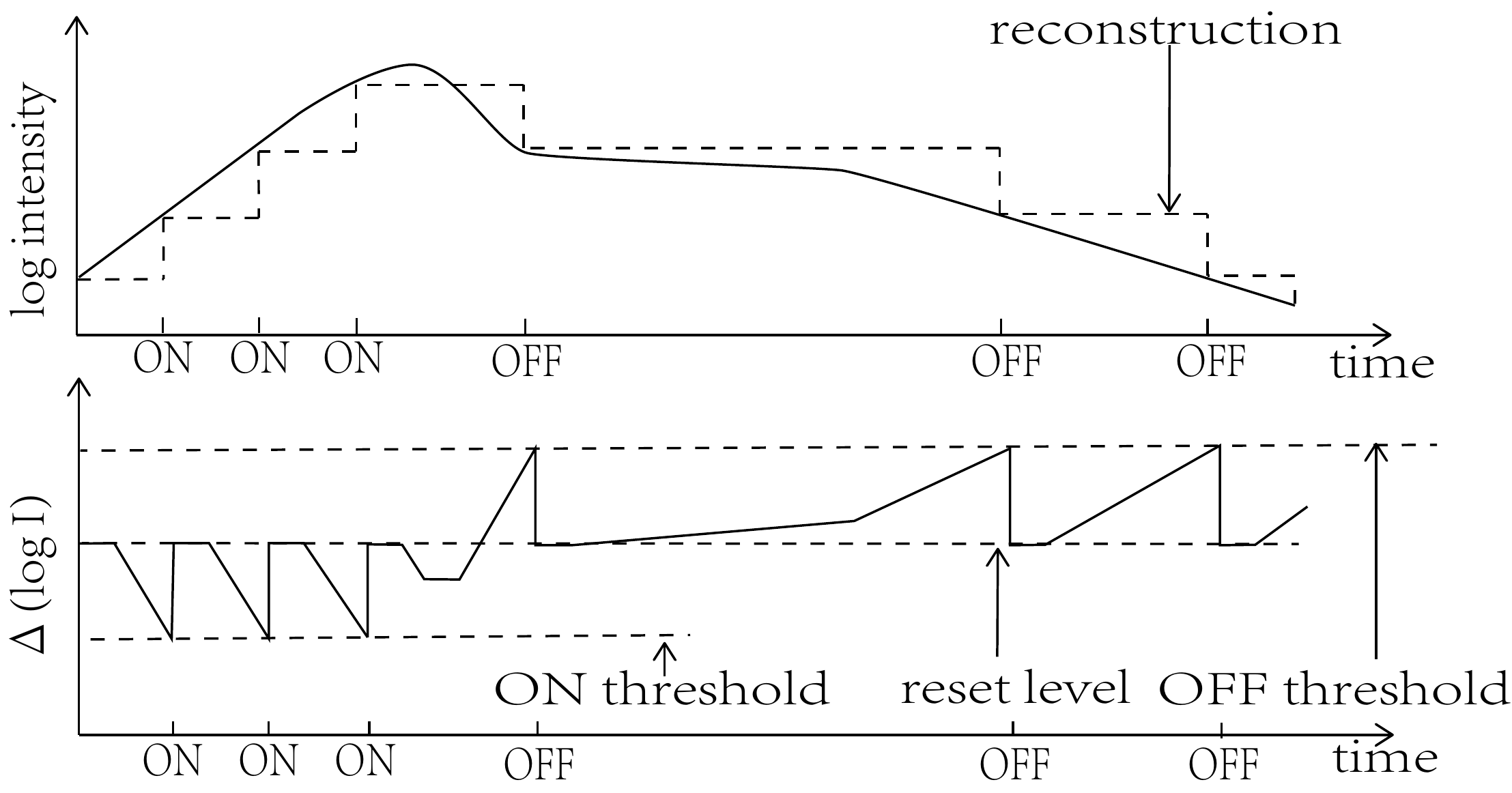}\label{fig_logA}}  
    \subfigure[Opposite light-dark processes.]{\includegraphics[width=0.495\linewidth]{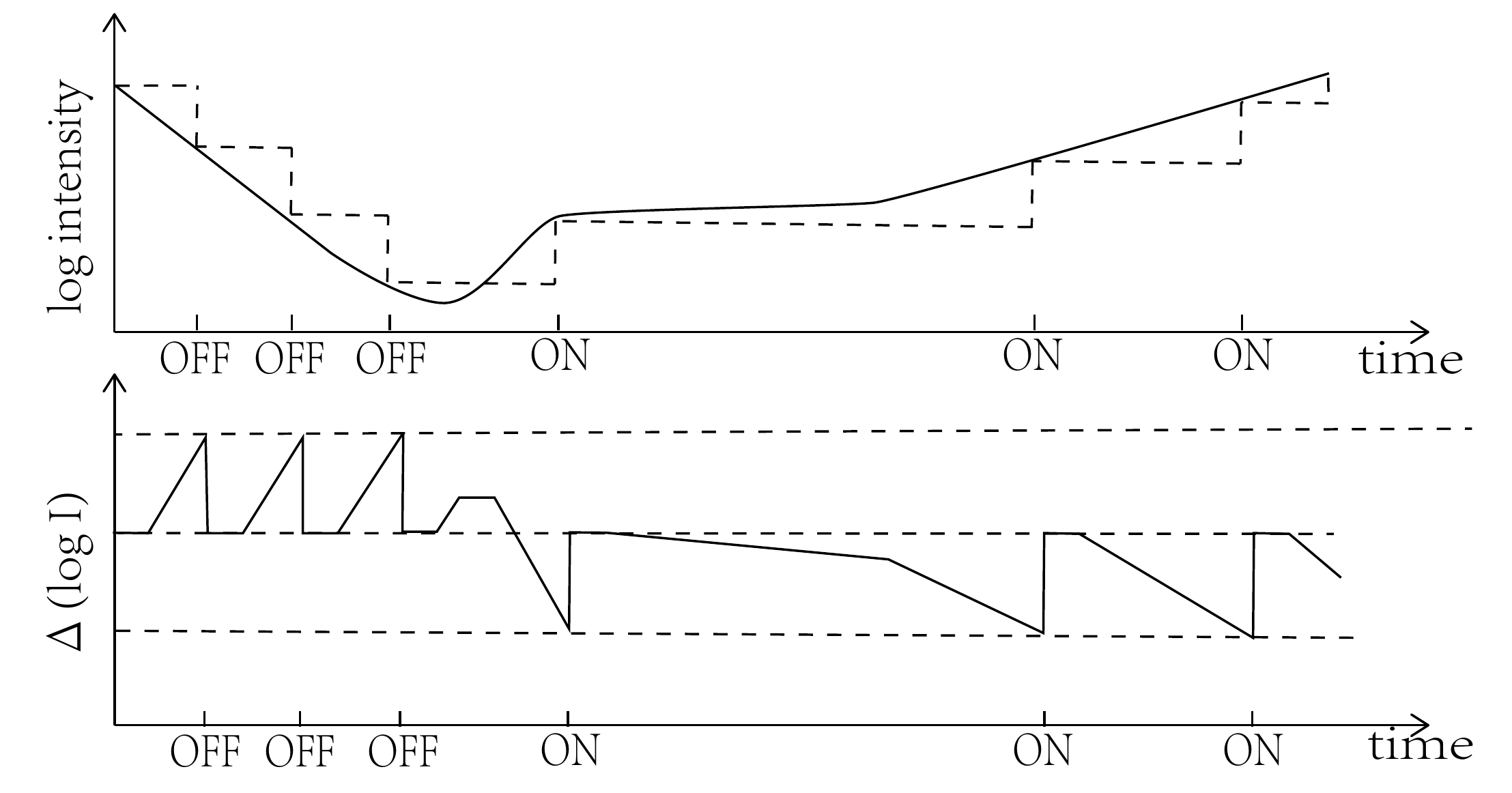}\label{fig_logB}}
    
    \subfigure[Inverted brightness variations.]{\includegraphics[width=0.495\linewidth]{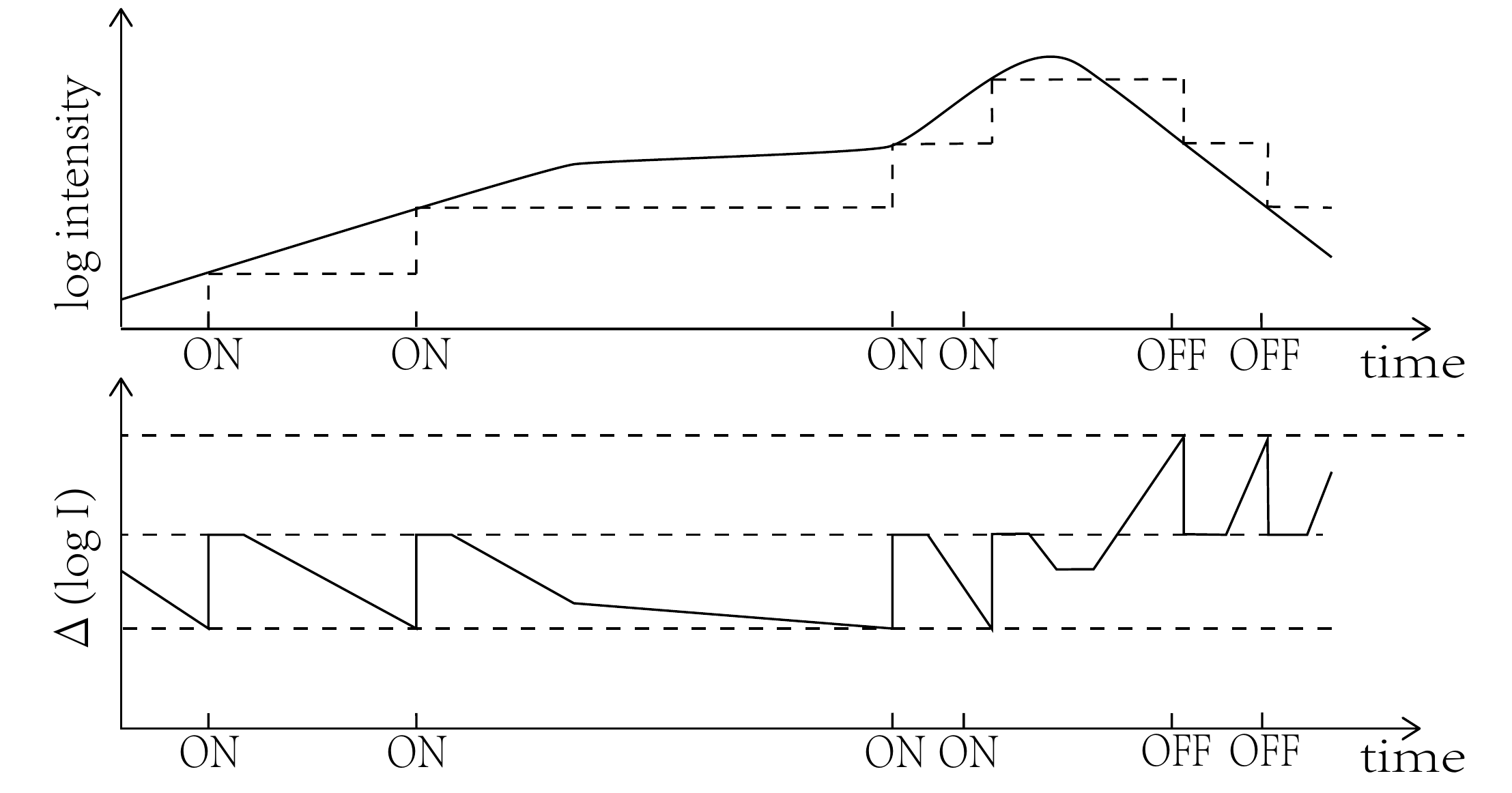}\label{fig_logC}}
    \subfigure[Delayed brightness variations.]{\includegraphics[width=0.495\linewidth]{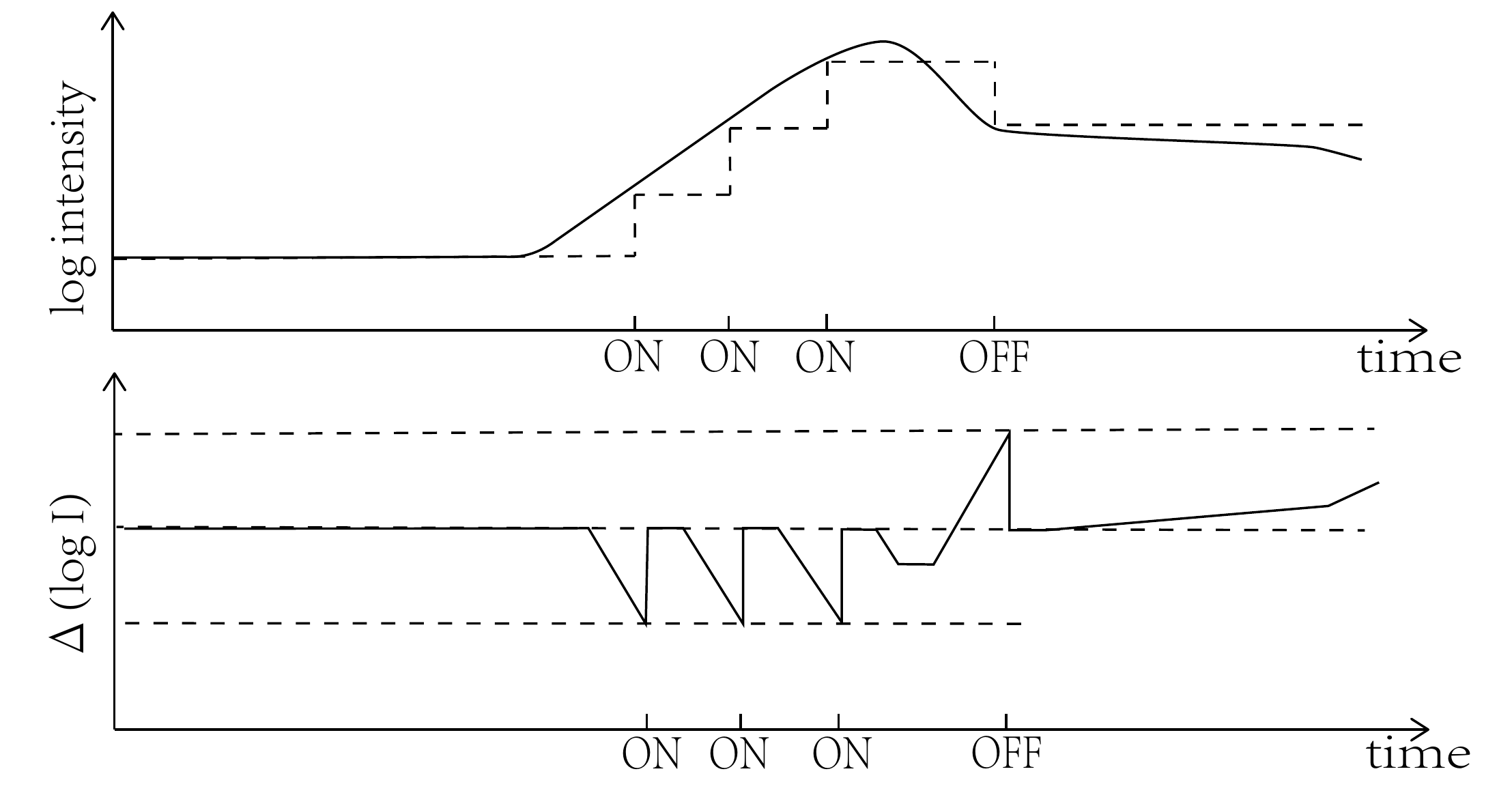}\label{fig_logD}}
    \caption{The relationship between logarithmic intensity and events generation for each pixel.
    (a) is the initial brightness and (b)(c)(d) is the changed brightness.}
    \label{fig_brightness} \vspace{-10pt}
\end{figure}

This work stems from three observations. 
First, the mammalian brain can make correct judgments 
based on fragments of time and space, 
while SNNs are less robust. 
In addition, it can be known from the event generation model 
that events asynchronously record brightness variances 
in microseconds. Thus, even slight brightness variances can
disrupt the time or polarity of events, 
leading to cascading changes in subsequent events. 
As shown in Figures~\ref{fig_logB}, ~\ref{fig_logC}, and ~\ref{fig_logD}, 
microsecond-level fluctuations in brightness can 
disrupt the spatiotemporal position and even the polarity of event fragments. 
Furthermore, sensor noise not only discards events~\cite{GuSHY21}, 
but may also disrupt the spatiotemporal location of event fragments.
For example, jitter in the step response may introduce 
sensor delays~\cite{DVSCamera}, 
resulting in delayed brightness variances. 
The relative mismatch between the reset level 
of the differential circuit and 
the comparator threshold may lead to a mismatch 
in the dynamic threshold $C$~\cite{DVSCamera}, 
causing the effect shown in Fig.~\ref{fig_logD}. 
These observations inspire us to improve the robustness 
of SNNs by artificially 
generating event spatiotemporal fragments.

\subsection{Event SpatioTemporal Fragment Augmentation}
To obtain the transformed event spatiotemporal fragments, 
the ESTF method is divided into two parts, 
inverting event fragments in spatiotemporal or polarity domains~(ISTP),
and drifting event fragments in spatiotemporal domain~(DST).

\textbf{ISTP} refers to inverting certain 
event fragments $\mathcal{E}_c$ on spatiotemporal or polarity domains $d$.
The transformed event field $S_{ISTP}$ can be
formulated as:
\begin{equation}\label{eventInvert}
    \begin{aligned}
        S^{\mathcal{E}_c}_{ISTP}(X, p, t)&=\sum_{e_{k} \in \mathcal{E}_c} (\delta \left(d + d_{k} - R_d\right) \prod_{i \in D\{d\}} \delta \left(i - i_k \right) )
    \end{aligned}
\end{equation}
where $d \in D$, $R_d$ represents the resolution of the domain $d$.
Event fragments are inverted from $d_k$ to their symmetrical 
positions $R_d - d_k$ in the event stream. 
When $d$ is the time domain $t$, $R_d$ represents the 
largest timestamp, when $d$ is the polarity $p$, 
$R_d$ represents 1, and when $d$ is the space coordinate $X$,
$R_d$ represents the image resolution.
$D\{d\}$ is the complement of $d$ with respect to $D$.

\begin{table*}[t]
    \centering 
    \caption{Performance of ESTF method and the SOTAs on CIFAR10-DVS, N-Caltech101 and N-CARS datasets. 
    }\label{tab_all_acc}
    \begin{tabular}{ccccccc}
    \toprule
    Network &Method& Reference & Model &CIFAR10-DVS & N-Caltech101 & N-CARS \\ \midrule
    \multirow{8}{*}{CNNs-based} & RG-CNNs~\cite{BiCABA20}    & TIP 2020 & Graph-CNN &54.00 &61.70 &91.40\\
    &  SlideGCN~\cite{LiZYZCB021}    & ICCV 2021 & Graph-CNN &68.00 &76.10 &93.10\\
    &  EventDrop~\cite{GuSHY21}    & IJCAI 2021& Mobilenet-V2 & - &87.14&94.55 \\ 
    &  ECSNet~\cite{ECSNet}    & T-CSVT 2022 & LEE$\rightarrow$MA &72.70  &69.30& 94.60\\
    & EV-VGCNN~\cite{deng2022voxel} &CVPR 2022 &EV-VGCNN& 67.00 &74.80&95.30 \\
    & NDA~\cite{li2022neuromorphic}    & ECCV 2022&Resnet-34&  -  &81.20&95.50\\
    \cmidrule(l){2-7}
    & Ours(w/o. ESTF) &-& Resnet-34& 77.15 & 90.86 & 93.80 \\
    & Ours(w/. ESTF) &-&Resnet-34& \textbf{83.12}\textcolor{blue}{$_{+5.97}$}& \textbf{93.15}\textcolor{blue}{$_{+2.29}$} &\textbf{96.00}\textcolor{blue}{$_{+2.20}$}\\ 
    \midrule 
    \multirow{10}{*}{SNNs-based} & HATS\cite{Sironi2018}& CVPR 2018 & HATS-SVM & 52.40 & 64.20 & 81.0 \\
    &Dart\cite{RameshYOTZX2020} & TPAMI 2020 & SPM-SVM & 65.80 & 66.80 & -  \\ 
    &STBP~\cite{zheng2021going} & AAAI 2021 & Resnet-19 &67.80 & -& -\\ 
    &Dspike~\cite{li2021differentiable}    & NeurIPS 2021 & ResNet-18 &75.40 & -& -\\ 
    &AutoSNN~\cite{na2022autosnn}    & ICML 2022& - &72.50 & -& -\\
    &RecDis~\cite{guo2022recdis}    & CVPR 2022& Resnet-19 &72.42& -& -\\
    &DSR~\cite{meng2022training}    & CVPR 2022& VGG-11 &77.27& -& -\\ 
    &NDA~\cite{li2022neuromorphic}    & ECCV 2022& VGG-11 & 81.70& 78.20 &90.1 \\ \cmidrule(l){2-7}
    & Ours(w/o. ESTF)                 & - &VGG-9 & 67.70& 67.32 &91.65\\ 
    & Ours(w/. ESTF)                 & - &VGG-9 & \textbf{83.90}\textcolor{blue}{$_{+16.20}$}& \textbf{80.07}\textcolor{blue}{$_{+12.75}$} &\textbf{95.64}\textcolor{blue}{$_{+3.99}$}\\ 
    \bottomrule
    \end{tabular}
\end{table*}

\textbf{DST} drifts events $\mathcal{E}_c$ 
a certain distance 
on spatiotemporal domain, which can be represented 
as a convolution kernel $k_{DST}$:
\begin{equation}\label{driftKernel}
  k_{DST}(X, p, t) = \delta \left(d - r_d \right) \prod_{i \in D\{d\}} \delta \left( i \right)
\end{equation}
where $d \in \{X, t\}$, $r_d$ represents the moving distance in the domain $d$. 
The transformed event field $S_{DST}$ can be obtained by 
convolving $k_{DST}$ with the event field $S$: 
\begin{equation}\label{eventDrift}
    \begin{aligned}
        S^{\mathcal{E}_c}_{DST}(X, p, t) &= \left(k_{DST} * S \right)(X, p, t) \\
        &= \sum_{e_{k} \in \mathcal{E}_c}k_{DST}(X-X_k,p-p_k, t-t_k) \\
  &=\sum_{e_{k} \in \mathcal{E}_c} (\delta \left(d - d_k - r_d\right) \prod_{i \in D\{d\}} \delta \left( i-i_k \right) )\\  
  \end{aligned}
\end{equation} 
where $d \in \{X, t\}$ and $\mathcal{E}_c$ represents 
events to be drifted. 
DST drifts fragments $\mathcal{E}_c$  
a distance $r$ in the domain $d$, and the part 
beyond borders will be discarded. 
It is worth noting that drifted samples of different scales are 
generated by different distances $r$, 
which obey a uniform distribution 
specified by a hyperparameter.
In addition, the event field $S_{ESTF}$ that applies both 
ISTP and DST can be obtained by convolving $k_{DST}$ and $S_{ISTP}$:
\begin{equation}\label{DriftR}
    S^{\mathcal{E}_c}_{ESTF}(X, p, t) = \left(k_{DST} * S^{\mathcal{E}_c}_{ISTP}\right)(X, p, t)
\end{equation}

The pseudocode of ESTF is given by Algorithm~\ref{alg_ESTF}.

\begin{algorithm}[tb]
    \caption{ESTF Procedure}\label{alg_ESTF}
    \textbf{Input}: {Original events $S=\left\{e_{i} \right\}_{i=1}^{N}$}\\
    \textbf{Parameter}: {Fragments ratio $c_{istp}$ and $c_{dst}$; Domain: $d_{istp}$ and $d_{dst}$; 
    Drift ratio: $r$; Resolution: $R_d$.}\\
    \textbf{Output}: Transformed event fragments $\mathcal{S}_{ESTF}$ 
    
    \begin{algorithmic}[1] 
      \STATE $begin \leftarrow Rand(0,1-c_{istp})$
      \STATE $\mathcal{E}_i \leftarrow \{e_i \}_{i \in [N*begin, N*(begin+c)]}$
      \STATE $S_{ISTP} = Eq 4(\mathcal{E}_i,d_{istp},R_d) + \mathcal{E}\{\mathcal{E}_i\}$      
      \STATE $begin \leftarrow Rand(0,1-c_{dst})$
      \STATE $\hat{r} \leftarrow Randint [-r*R_d, r*R_d]$\;
      \STATE $\mathcal{E}_d \leftarrow \{e_i \}_{i \in [N*begin, N*(begin+c)]}$
      \STATE $S_{ESTF} = Eq 7(S^{\mathcal{E}_d}_{ISTP},d_{dst},\hat{r}) + \mathcal{E}\{\mathcal{E}_d\}$      
      \STATE \textbf{return} $S_{ESTF}$
    \end{algorithmic}
\end{algorithm}

\section{Experiments}




\subsection{Implementation}\label{sec_implement}
Extensive experiments are performed to demonstrate 
the superiority of our ESTF method on 
the prevailing neuromorphic datasets, 
including CIFAR10-DVS~\cite{Li2017}, 
N-Caltech101~\cite{N101}, N-CARS~\cite{Sironi2018} datasets.
For the convenience of comparison, 
the model with the same parameters without ESTF 
is used as the baseline. 
STBP~\cite{zheng2021going} method is used to train SNNs, and 
other parameters mainly refer to NDA~\cite{li2022neuromorphic}.
For example, the initial learning rate is $1e-3$, 
the neuron threshold and leakage coefficient are 1 and 0.5, 
respectively. 
Furthermore, we also evaluate the performance of 
ESTF on CNNs, a popular network for processing neuromorphic data. 
The EST~\cite{EST} representation method 
is used to convert neuromorphic data 
into a form compatible with CNNs. 
Adam optimizer is used with an initial learning rate of $1e-4$.

\vspace{-10pt}
\subsection{Compared with SOTAs}
ESTF outperforms previous SOTAs
on three prevailing and challenging neuromorphic datasets.
As shown in Tab.~\ref{tab_all_acc}, 
ESTF achieves substantial improvements on \textbf{SNNs} 
through spatiotemporal fragments. 
In particular, ESTF achieves a \textbf{16.2\%} improvement 
on the CIFAR10-DVS dataset, 
reaching a state-of-the-art accuracy of \textbf{83.9\%}.
In addition,
ESTF also has significant improvements over \textbf{CNNs}.
Since ESTF is orthogonal to most training algorithms, 
it can provide a better baseline and improve the 
performance of existing models.

\begin{table}[t]
    \centering
    \caption{Performance of ESTF compared with other event data augmentations 
    on the N-Caltech101 dataset.}\label{tab_augmentation}
    \begin{tabular}{cccc}
        \toprule 
        \multirow{3}{*}{Methods} & \multicolumn{3}{c}{Accuracy(\%)} \\ \cmidrule(l){2-4}
                            & VGG-9   & VGG-19         & Resnet-34      \\  
                            & (SNNs)   & (ANNs)        & (ANNs)        \\ \midrule
        Baseline            & 67.32  & 87.30   &  90.86   \\ 
        EventDrop           & 68.69  & 88.31   &  92.06   \\ 
        NDA$^*$             & 77.44   & 88.84   &  92.42 \\ 
        ESTF                & \textbf{80.07}\textcolor{blue}{$_{+12.75}$}  & \textbf{91.29}\textcolor{blue}{$_{+3.99}$}   &  \textbf{93.15}\textcolor{blue}{$_{+2.29}$}  \\ \bottomrule
    \end{tabular}
    \vspace{-10pt}
\end{table}
\begin{figure}[b]
    \vspace{-10pt}
    \subfigure[ISTP and DST.]{\includegraphics[width=0.45\linewidth]{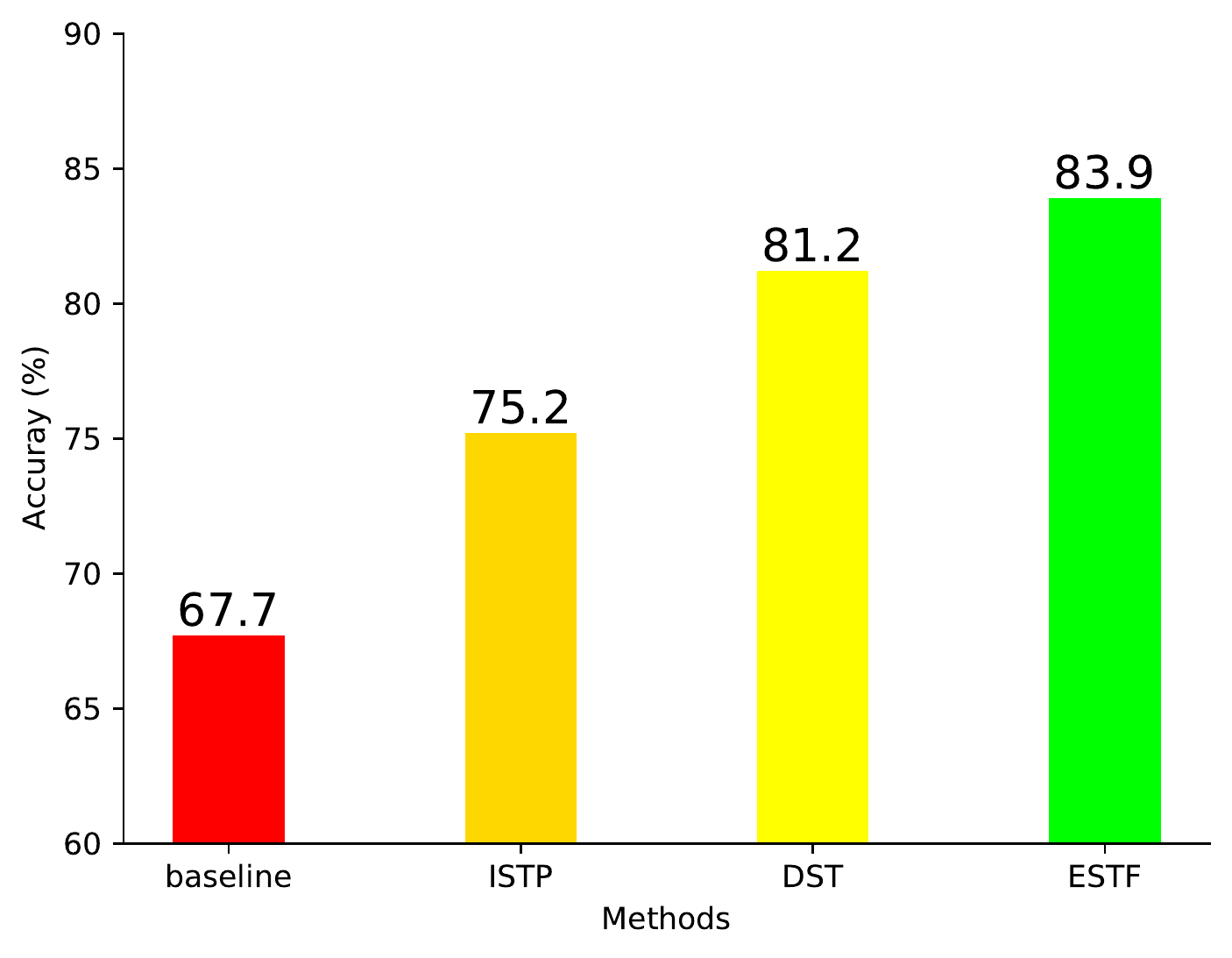}\label{fig_ST}}  
    \subfigure[Different domains.]{\includegraphics[width=0.45\linewidth]{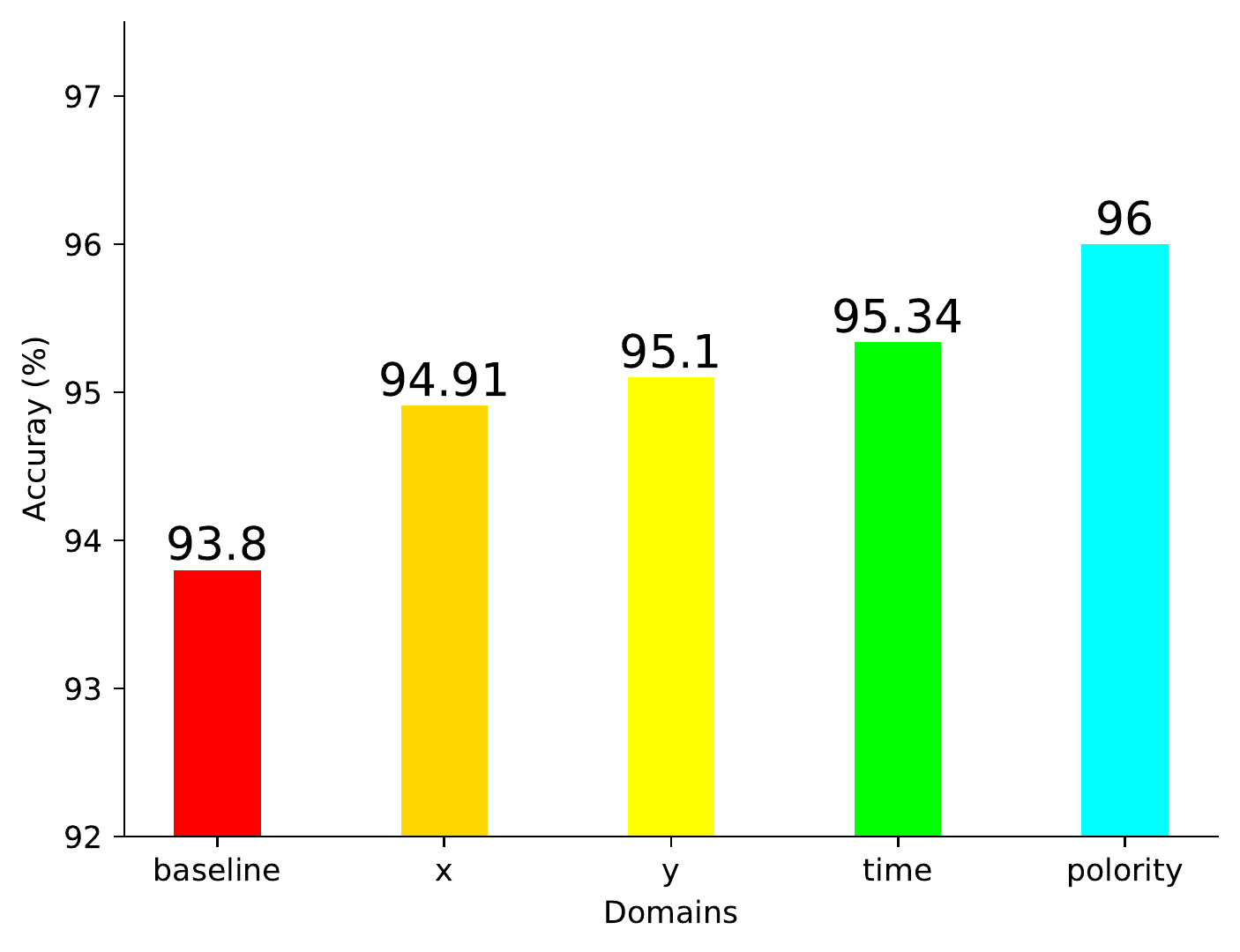}\label{fig_domains}}
    \caption{Ablation studies on ESTF.}\label{fig_ablation}
\end{figure} 
\vspace{-10pt}
\subsection{Compared with Event Data Augmentations}
As shown in Tab.~\ref{tab_augmentation}, 
ESTF is compared with other reproduced event data augmentations 
at the same baseline. 
EventDrop~\cite{GuSHY21} is similar to Cutout on event data
and achieves remarkable success on CNNs. 
NDA$^*$ represents applying the two naive geometric transformations 
flipping and translation on event data.
It turns out that ESTF outperforms EventDrop 
and NDA$^*$ in multiple networks, 
covering both SNNs and ANNs. 
It is worth noting that 
the performance of EventDrop in SNNs may lag far behind ESTF
because deletion destroys the continuity of neuromorphic data
and may cause the ``dead neuron" problem. 
In addition, ESTF is more stable and more effective than
geometric transformations.
It illustrates the superiority of spatiotemporal fragment
augmentation.

\subsection{Ablation Studies on ESTF}
\textbf{Performance of ISTP and DST}.
As shown in Fig.~\ref{fig_ST}, 
the performance of ISTP and DST are evaluated on the DVS-CIFAR10 dataset. 
It turns out that both ISTP and DST have significant improvements. 
Notably, the improvement of DST is more pronounced
since drifting events are more common than inverting events.

\textbf{Performance of Different Domains}.
As shown in Fig.~\ref{fig_domains}, 
the performance of ESTF in different domains is evaluated 
on the N-CARS dataset.
It turns out that ESTF is effective in the time, 
spatial, and polarity domains. 
It is worth noting that the ESTF performs better 
in the temporal and polarity domains 
than in the spatial domain, 
which indicates that the event spatiotemporal fragments 
are critical to improving the robustness of the model.


\vspace{-10pt}
\subsection{Analysis of ESTF} 
\begin{figure}[t]
    \subfigure[Inverted brightness variations.]{\includegraphics[width=0.45\linewidth]{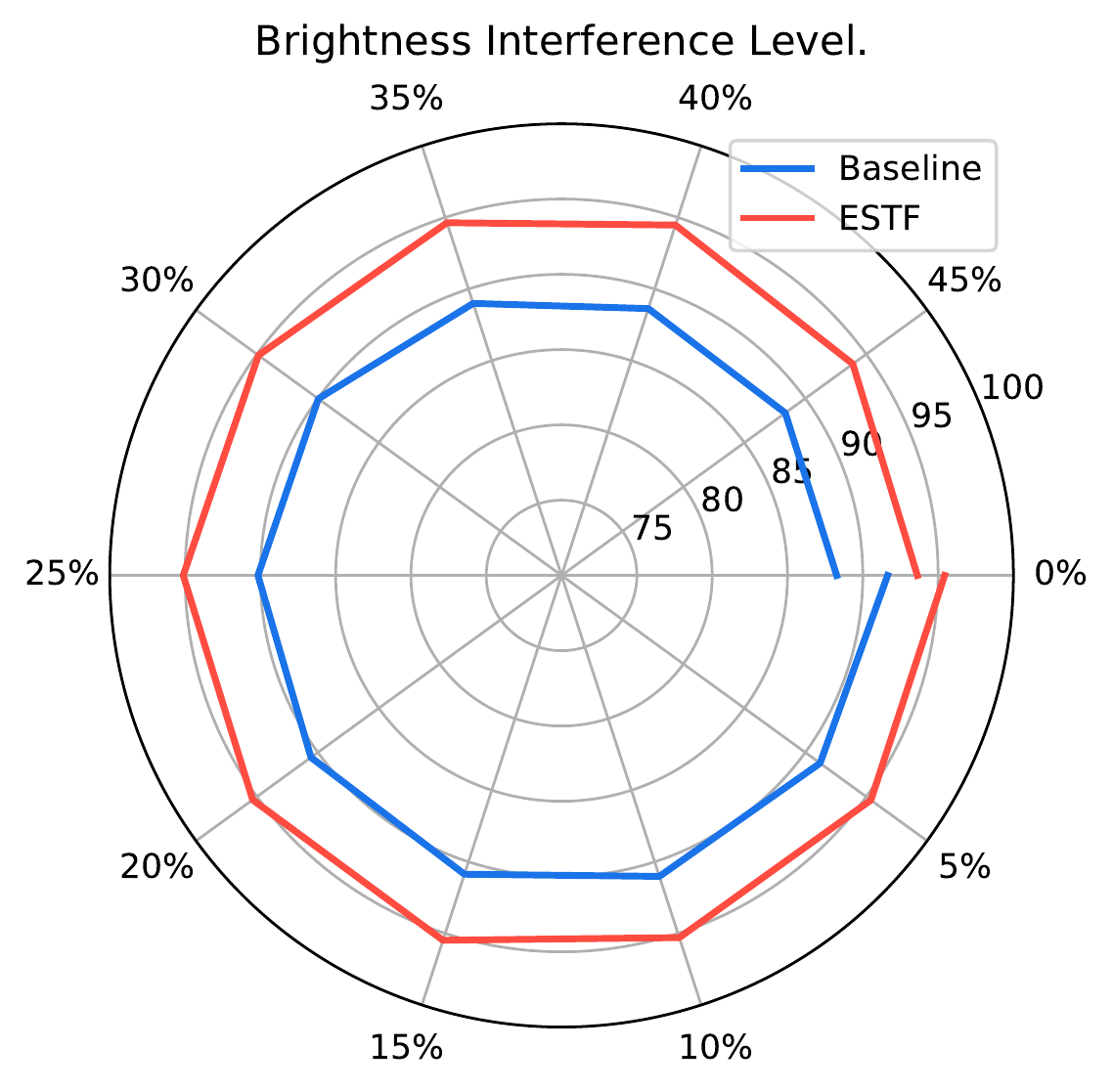}\label{fig_LevelB}}
    \subfigure[Delayed brightness variations.]{\includegraphics[width=0.45\linewidth]{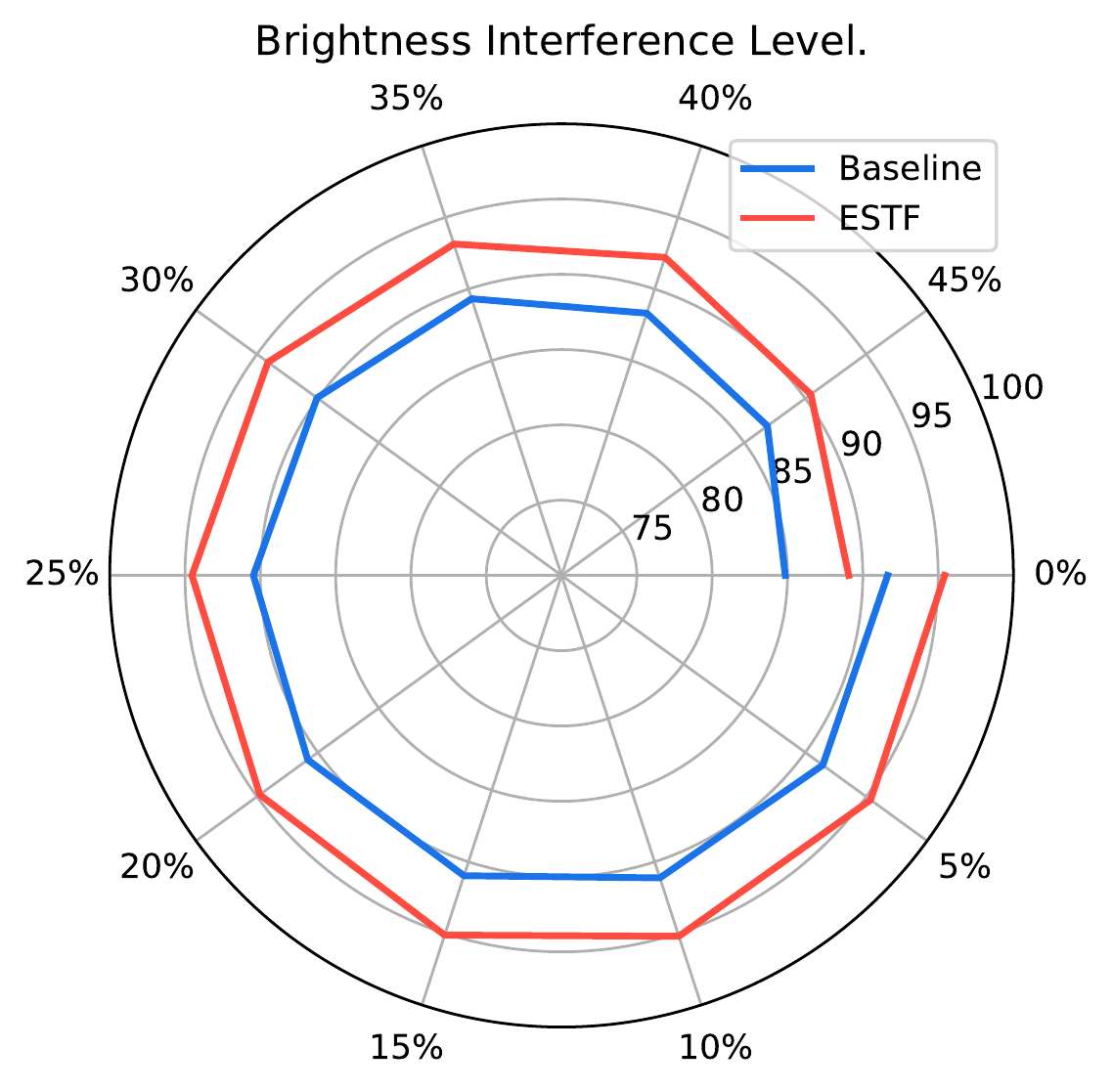}\label{fig_LevelA}}  
    \caption{Performance under different brightness interferences.}\label{fig_variations}
\end{figure} 
There are at least three insightful reasons for the 
superior performance of the event spatiotemporal augmentation (ESTF).
Intuitively, ESTF transforms in the time, space, 
and polarity domains, 
while NDA only transforms in space. 
Fig~\ref{fig_domains} shows the performance of ESTF 
in different domains. 
It is worth noting that the effects of transformations 
in the time and polarity domains are significant. 
Considering that the performance of 
NDA$^*$ geometric transformations 
is already close to SOTAs, 
there is still more than 2\% improvement in ESTF, 
which is quite impressive, as shown in Tab.~\ref{tab_augmentation}.
In addition, ESTF preserves the continuity 
of neuromorphic data through spatiotemporal fragments, 
maintaining the original local features. 
The comparison with EventDrop in Tab.~\ref{tab_augmentation} 
illustrates the importance of continuity.
Furthermore, the transformations of ISTP and DST 
in time and polarity domains essentially 
improve the robustness of SNNs to brightness variances. 
Note that event samples may only last 100ms~(N-CARS), 
and slight brightness fluctuations can bring brightness variances similar to 
Fig.~\ref{fig_brightness}.
As shown in Fig.~\ref{fig_variations}, 
we further evaluate the robustness of SNNs against brightness variances. 
It turns out that the model with ESTF is significantly more robust 
under various degrees of brightness interference.

\vspace{-5pt}
\section{Conclusion}    
\vspace{-5pt}
In this paper, we propose event spatiotemporal fragment augmentation. 
It preserves the continuity of the data, 
transforms events in the spatiotemporal and polarity domains, 
and simulates the influence of brightness variances on neuromorphic data. 
It makes up for the omission of spatiotemporal fragment augmentation in the past,
resulting in superior improvements and SOTA performance 
on prevailing neuromorphic datasets.

%


\newpage
\bibliographystyle{IEEEbib}
\bibliography{icassp23}

\end{document}